\documentclass[draft]{agujournal2019}
\usepackage{url} 
\usepackage{lineno}
\usepackage[inline]{trackchanges} 
\usepackage{soul}
\usepackage{multirow}
\usepackage{array} 

\draftfalse

\journalname{Journal of Advances in Modeling Earth Systems (JAMES)}

\begin{document}

%

\title{Intelligent model for offshore China sea fog forecasting}
%

%
%




\authors{Yanfei Xiang\affil{1}, Qinghong Zhang\affil{2}, Mingqing Wang\affil{1}, Ruixue Xia\affil{1}, Yang Kong\affil{3,4}, Xiaomeng Huang\affil{1}} 

\affiliation{1}{Department of Earth System Science, Ministry of Education Key Laboratory for Earth System Modeling, Institute for Global Change Studies,Tsinghua University, Beijing 100084, China}
\affiliation{2}{Department of Atmospheric and Oceanic Sciences, School of Physics, Peking University, Beijing 100871, China}
\affiliation{3}{
Ningbo Meteorological Bureau, Ningbo, Zhejiang 315012, China}
\affiliation{4}{Ningbo Meteorological Disaster Warning Center, Ningbo, Zhejiang 315012, China}

\correspondingauthor{Xiaomeng Huang}{hxm@mail.tsinghua.edu.cn}



\begin{keypoints}
\item We introduce a time-lagged correlation analysis (TLCA) method capable of evaluating the time-lagged influence of meteorological predictors related to fog. Essential predictors identified by the TLCA method are used to investigate the mechanism of sea fog.
\item We build a unified forecasting method that allows forecasting for multiple stations. To tackle the data imbalance between fog and fog-free events during ML-based model training, we utilize focal loss and an ensemble learning strategy to improve the prediction skills.
\item Our ML-based method outperforms two traditional methods, including the WRF-NMM model and the NOAA FSL method, in predicting sea fog (horizontal visibility $\le$ 1 km) with 60-hours lead time, by improving probability of detection (POD) while reducing false alarms ratio (FAR).
\end{keypoints}

%
%

%
%

\begin{abstract}
Accurate and timely prediction of sea fog is very important for effectively managing maritime and coastal economic activities. Given the intricate nature and inherent variability of sea fog, traditional numerical and statistical forecasting methods are often proven inadequate. This study aims to develop an advanced sea fog forecasting method embedded in a numerical weather prediction model using the Yangtze River Estuary (YRE) coastal area as a case study. Prior to training our machine learning model, we employ a time-lagged correlation analysis technique to identify key predictors and decipher the underlying mechanisms driving sea fog occurrence. In addition, we implement ensemble learning and a focal loss function to address the issue of imbalanced data, thereby enhancing the predictive ability of our model. To verify the accuracy of our method, we evaluate its performance using a comprehensive dataset spanning one year, which encompasses both weather station observations and historical forecasts. Remarkably, our machine learning-based approach surpasses the predictive performance of two conventional methods, the weather research and forecasting nonhydrostatic mesoscale model (WRF-NMM) and the algorithm developed by the National Oceanic and Atmospheric Administration (NOAA) Forecast Systems Laboratory (FSL). Specifically, in regard to predicting sea fog with a visibility of less than or equal to 1 km with a lead time of 60 hours, our methodology achieves superior results by increasing the probability of detection (POD) while simultaneously reducing the false alarm ratio (FAR).

\textbf{Keywords}: sea fog, weather forecast, key predictors, machine learning, data imbalance
\end{abstract}

\section*{Plain Language Summary}

Accurate and timely prediction of sea fog is very important for managing activities related to maritime and coastal economies. However, traditional methods of forecasting using numbers and statistics often struggle because sea fog is complex and unpredictable. In this study, we aimed to develop a smart machine learning method to predict sea fog. We focused on the Yangtze River Estuary (YRE) coastal area as an example. Before training our model, we used a technique named time-lagged correlation analysis (TLCA) to determine which factors are important for causing sea fog. We also used ensemble learning and a special method named the focal loss function to handle situations where there is an imbalance between the fog and fog-free data. To test how accurate our method is, we used much data from weather stations and historical forecasts over the course of one year. Our machine learning approach performed better than two other methods—the weather research and forecasting nonhydrostatic mesoscale model (WRF-NMM) and the method developed by the National Oceanic and Atmospheric Administration (NOAA) Forecast Systems Laboratory (FSL)—in predicting sea fog. Specifically, in regard to predicting sea fog 60 hours in advance with a visibility of 1 kilometer or less, our method achieved better results by increasing the chances of detecting sea fog while reducing false alarms.


\section{Introduction}

The cooling effects of water vapor condensation near the Earth's surface give rise to sea fog, a powerful meteorological phenomenon that occurs in coastal areas and at sea. This process leads to the formation of numerous water droplets or ice crystals, resulting in a significant reduction in air visibility to less than 1 km \cite{wang1983a}. The impact of sea fog can be comparable to that of tornadoes or hurricanes, affecting offshore fishing, shipping, platform operations, coastal air quality, and road transportation \cite{gultepe2007b}. However, predicting sea fog is challenging due to the intricate interplay between synoptic (at a larger scale), mesoscale (at a regional scale), and local factors \cite{gultepe2007a, lewis2004b, fabbian2007a, kora2014a}.

Operational sea fog forecasting currently employs three main methods: dynamical methods, statistical methods, and statistical–dynamical hybrid methods. Dynamical models, such as numerical weather prediction (NWP) models, simulate atmospheric evolution by solving dynamic and thermodynamic equations. However, predicting sea fog with NWP models is challenging due to incomplete initial conditions \cite{gao2007a, wang2014a},  uncertain parameterization schemes \cite{gultepe2006a, zhong2015a}, and coarse resolution \cite{gultepe2007a, xu2015a}. Statistical models, on the other hand, approximate relationships between observed visibility-related factors without explicitly considering the underlying physical processes. Regression analysis has been used to predict visibility using meteorological factors \cite{dewi2020a, fabbian2007a, yu2007a}. Some studies, such as SW99 \cite{stoelinga1999a}, the US National Oceanic and Atmospheric Administration (NOAA) Forecast Systems Laboratory (FSL) algorithm \cite{doran1999a}, and the US Air Force Weather Agency (AFWA) diagnostic procedure in the Weather Research and Forecast model (WRF) \cite{creighton2014a}, have incorporated microscopic predictors into empirical equations. Furthermore, statistical–dynamical hybrid models combine atmospheric variables from dynamical models with statistical approaches \cite{hu2014a, lewis2004a, bang2008a}. However, these models often select predictors based on correlation analysis without considering time lag effects. Machine learning (ML) is an innovative technique that can complement existing methods and enhance our understanding of Earth systems \cite{reichstein2019a}. ML models have the ability to learn from extensive datasets, enabling them to uncover underlying physical principles and make accurate predictions. Consequently, they have proven to be valuable tools in weather forecasting applications \cite{shi2015a, shi2017a, ravuri2021a}. Researchers have successfully utilized ML models, such as decision trees \cite{lewis2004a, huang2011a} and artificial neural networks (ANNs), to predict fog \cite{fabbian2007a, colabone2015a}. The findings of these studies suggest that ML models outperform traditional approaches in terms of predictive ability. This improvement can be attributed primarily to the significant nonlinear fitting capacity of ML models, which allows for the extraction of essential physical information from data.

In this study, we present a novel approach for predicting sea fog by utilizing the output from a numerical weather prediction (NWP) model. We select the Yangtze River Estuary (YRE) as a case study due to its significance as an economic hub in China. Accidents related to marine transportation and operations have frequently occurred in this region due to the occurrence of fog \cite{gu1993a, wu2014a, shi2016a}. Accurate prediction of sea fog in the YRE can greatly benefit transportation and commercial activities. The following is a brief overview of our significant contributions:

\begin{enumerate}
\item \textbf{Time-lagged correlation analysis (TLCA)}: We introduce an innovative technique named TLCA, which enables us to assess the time-delayed impact of meteorological predictors associated with sea fog. By employing the TLCA method, we identify crucial predictors that offer insights into the intricate mechanisms of sea fog formation.
\item \textbf{Comprehensive forecasting approach}: We develop a comprehensive forecasting approach capable of generating predictions for multiple weather stations. To address the inherent data imbalance between fog and fog-free occurrences during machine learning model training, we employ the focal loss and utilize an ensemble learning strategy. These measures effectively enhance the predictive capabilities of our model.
\item \textbf{Superior performance in sea fog prediction}: The performance of our ML-based method surpasses the performance of two traditional methods—the WRF-NMM and the NOAA FSL method—in the domain of sea fog prediction. Specifically, our approach achieves superior outcomes in predicting sea fog events with a horizontal visibility of 1 kilometer or less up to 60 hours in advance. Notably, we improve the probability of detection (POD) while simultaneously reducing the false alarm ratio (FAR).
\end{enumerate}

\section{Materials and Methods}

\subsection{Study Area}

The study focuses on the Yangtze River Estuary (YRE), which includes sea, river, and land areas extending from 26.5°N to 33.5°N and from 117°E to 126°E. The YRE is in a subtropical monsoon climate zone with an annual precipitation of over 1000 mm, contributing abundant water resources to the atmosphere. Moreover, both industrial facilities on land and oceanic waves release significant amounts of condensation nuclei, such as dust and salt particles, into the atmosphere, facilitating the condensation of water vapor into fog droplets \cite{clarke2003a, quan2011a, gu1993a}. As a result, the YRE exhibits favorable conditions for the formation of sea fog.

\subsection{Data and Preprocessing}

\begin{figure}
\centering
\includegraphics[width=12cm]{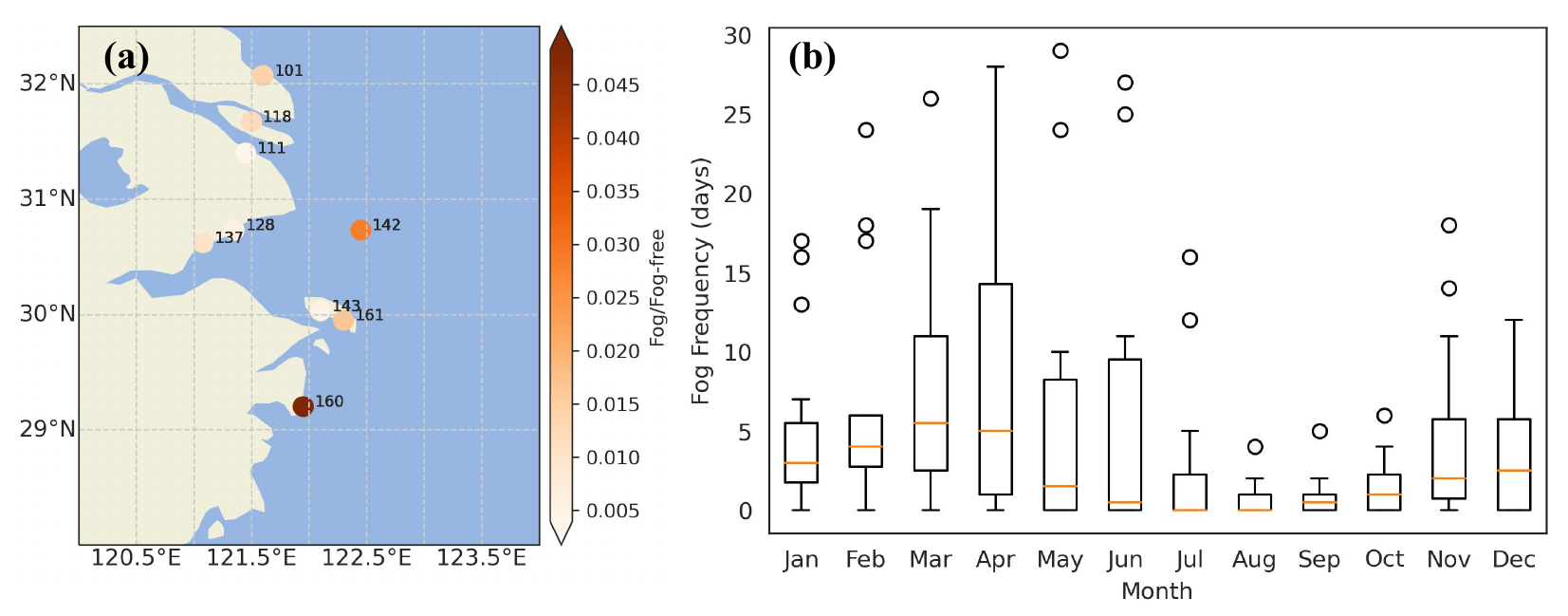}
\caption{(a) Geographical distribution and fog/fog-free sample ratio at coastal weather stations; and (b) temporal distribution of fog events in the YRE across 9 stations.}
\label{Figure 1}
\end{figure}

This study utilized observations of visibility conditions and numerical weather forecast data from 2014 to 2020. The observations were obtained from the Meteorological Information Comprehensive Analysis and Process System (MICAPS), which provides various weather measurements, including temperature, pressure, humidity, precipitation, visibility, current weather, and other meteorological variables. We can determine if there is fog based on two items: current weather and visibility. Meteorological measurements were recorded at 3-hour intervals. Figure \ref{Figure 1} (a) shows the spatial distribution of weather stations in the YRE, where the color indicates the ratio of fog to fog-free events recorded at each station. Our findings indicate that fog is more likely to occur in the southern region of the YRE. Furthermore, we observed that fog tends to increase in March, peak in April, and decrease in June. Between August and October, there were fewer instances of fog events, while the frequency of fog events gradually increased from November to February, as depicted in Figure \ref{Figure 1}(b).

We utilized historical forecast data from the weather research and forecasting nonhydrostatic mesoscale model (WRF-NMM) provided by the US National Centers for Environmental Prediction (NCEP) \cite{janjic2003a}. The horizontal resolution of the forecast data is $8 \times8$ km , covering the entire Yangtze River Estuary (YRE) region. WRF-NMM produces hourly predictions for a 60-hour forecast period, launching twice daily at 00:00 UTC and 12:00 UTC. The forecast variables incorporated in the model include temperature, relative humidity, cloud cover, wind speed, precipitation, and more. We selected 26 variables (see Table \ref{Table A1}) and omitted certain nearly constant variables. To integrate the gridded forecast data with sparse observation data, we used a two-step preprocessing method. First, we synchronized the time zones of the two datasets to UTC+00:00. Second, we used the inverse distance weighted (IDW) approach for spatial matching \cite{yang2015a}. As a result, we generated a multivariable time series dataset as follows:

\begin{linenomath*}
\begin{equation}
 D=\{X_{N\times M\times T},Y_{N\times T}\} 
\end{equation}
\end{linenomath*}

where $X$ is a three-dimensional tensor of size $N\times M\times T$ (corresponding to $N$ samples of $M$ forecast variables of $T$ hours), and $Y$ is the observational data of size $N\times T$ (corresponding to $N$ samples of $T$ hours). A sample s from dataset D is represented as $s= \{ x_{M\times T},y_T \} \in D$. In this study, $N=1637335$, $M=27$, and $T=60$. Sea fog is a regional meteorological phenomenon with distinct causes in different places \cite{kora2014a}. Therefore, it is critical to examine the major factors that contribute to sea fog in the YRE.

\subsection{Time-lagged Correlation Analysis}

We introduced the time-lagged correlation analysis (TLCA) method to gain a deeper understanding of the crucial factors that influence fog formation and selected predictors for the fog forecasting method described in Section 2.4. Our TLCA method considers the time lags present in atmospheric time-series data, which have been noted in prior research \cite{kawale2012a}. Although there have been limited studies on the impact of time-lagged effects on feature selection \cite{colabone2015a, kim2020a}, identifying time-lagged linkages is a critical task. Therefore, examining time-lagged relationships could greatly enhance the accuracy of sea fog forecasting. We performed TLCA analyses on fog events that occurred between March and July, during which fog events are most frequent in the YRE, as shown in Figure \ref{Figure 1}(b). The details of the TLCA algorithm are described below.

\begin{enumerate}
\item We set the maximum lag time $\tau$ and select a subdataset from  $D$, which is denoted as $D_{\mathrm{TCCA}}:D_{\mathrm{TCCA}}=\{X_t,y_0\}\subseteq D,=0,-1,-2,\ldots,-\tau$. The observed visibility is denoted as $y_0$. The forecast variables before the observation time are denoted as $X_t$. In this work, we set $\tau=5$ because of the trade-off between the number of predictors and forecast accuracy.
\item By using the Pearson correlation coefficient, we determined how closely $y_0$ is related to each of the variables in $X_t$. The variables that do not pass the significance test were omitted from the analysis (the threshold for statistical significance was set to $\alpha=0.05$). 
\end{enumerate}

\begin{figure}
\centering
\includegraphics[width=12cm]{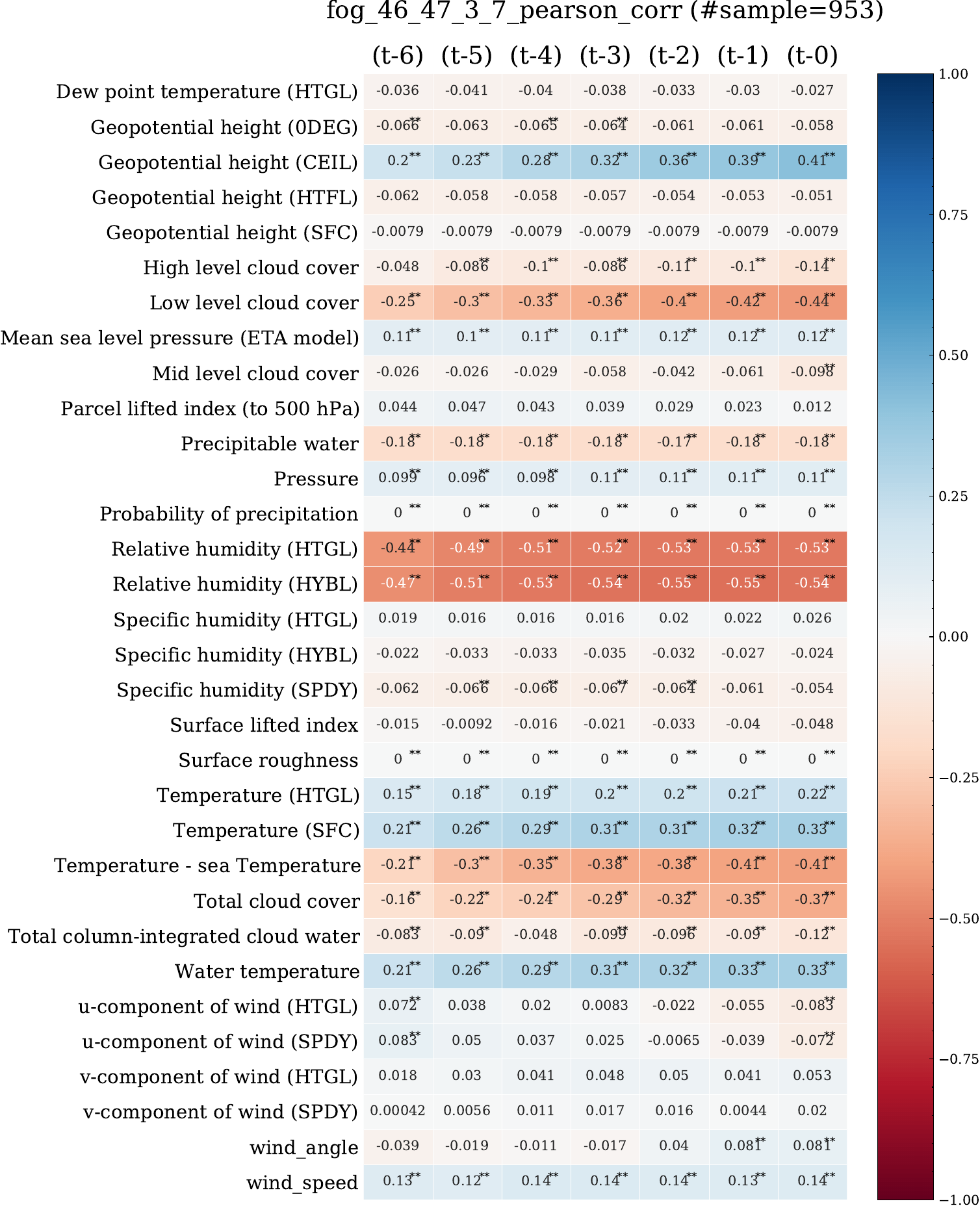}
\caption{Time-lagged correlations between visibility (fog events) and meteorological variables between March and July. Each row indicates the meteorological variables, and each column indicates the lag time.}
\label{Figure 2}
\end{figure}

\begin{figure}
\centering
\includegraphics[width=12cm]{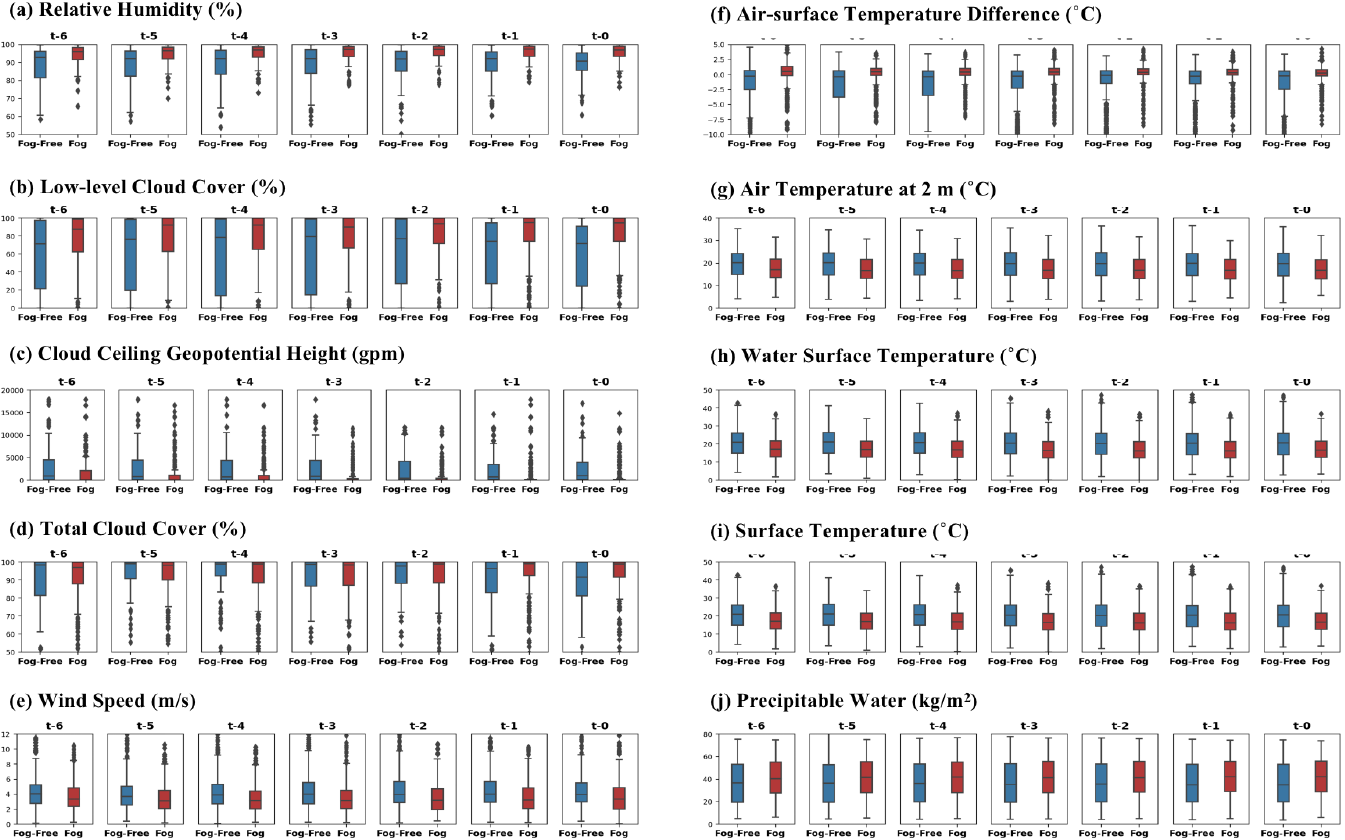}
\caption{Meteorological variables in fog and fog-free events across all stations between March to July.}
\label{Figure 3}
\end{figure}

We identified 10 meteorological variables that are highly correlated with visibility (see Figure \ref{Figure 2}), including relative humidity, low-level cloud cover, cloud ceiling geopotential height, total cloud cover, air–sea temperature difference, sea surface temperature, surface temperature, precipitable water, wind speed, and air temperature at 2 meters. All these variables have a significant correlation with a time lag. The formation of fog is heavily influenced by relative humidity. As relative humidity increases, the air can hold more water vapor, which increases the likelihood of fog formation \cite{syed2012a, ding2013a}. Typically, fog forms when the relative humidity reaches or exceeds 90\%, as shown in Figure \ref{Figure 3}(a). As the moist air near the surface cools to the dew point temperature, water vapor condenses into tiny droplets, forming a low-level cloud or fog \cite{ukhurebor2017a, bergot2005a, gultepe2007b}. The total cloud cover includes low-level clouds and is therefore strongly correlated with fog formation \cite{sedlar2021a, thompson2005a}. When the geopotential height of the cloud ceiling is low, it indicates a stable atmosphere and high moisture levels near the surface \cite{schweiger2008a, ma2018a}, which creates favorable conditions for fog formation \cite{roco2018a}, especially in areas with high relative humidity and temperatures close to the dew point. In coastal regions, a greater temperature difference between the air and the surface can increase the likelihood of fog formation. One of the most common types of fog associated with air–sea temperature differences is advection fog, which occurs when warm, moist air moves over a cooler surface such as water or land \cite{gao2007a, kawai2015a, yang2018a}. Precipitable water measures the total amount of water vapor in the atmosphere \cite{albano2020a}. High levels of precipitable water indicate a significant amount of moisture in the air, which can increase the likelihood of fog formation. Fog events often occur in high-pressure weather systems with calm winds \cite{guo2015a}. According to Figure \ref{Figure 3}(e), wind speeds below 5 m/s are favorable for fog formation in the YRE.

\begin{figure}
\centering
\includegraphics[width=12cm]{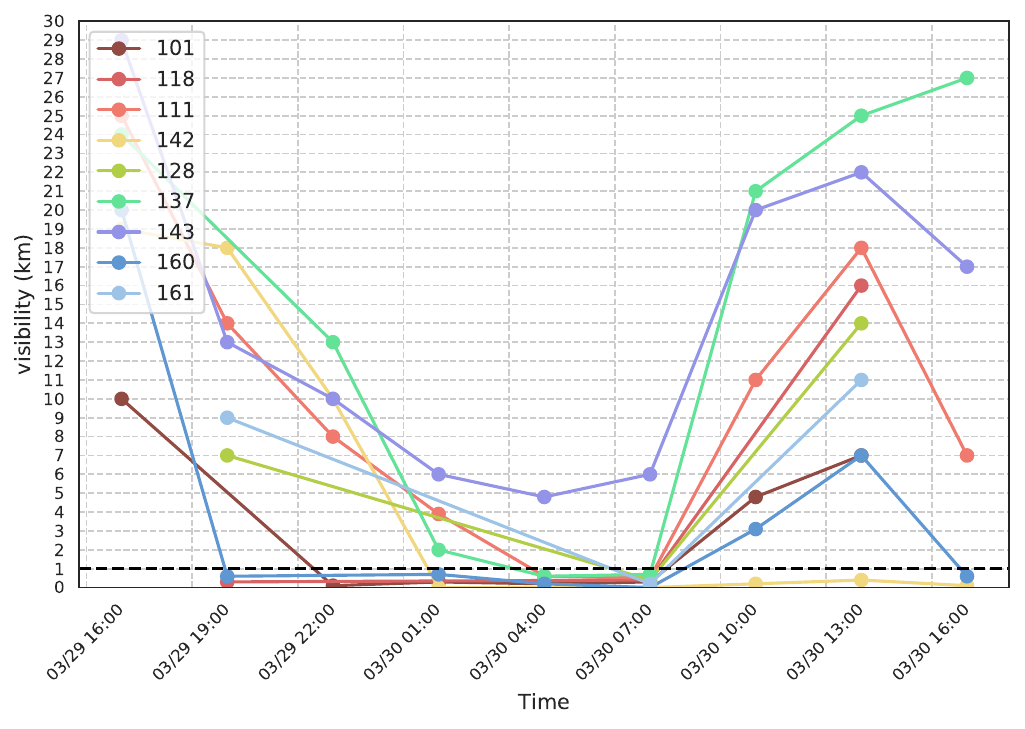}
\caption{Observed visibility of 9 stations in the YER from 17:00 on March 29th to 17:00 on March 30th, 2018.  The locations of the stations are shown in Figure \ref{Figure 1}. The black horizontal dashed line indicates a visibility threshold of 1 km. Some stations have missing records.}
\label{Figure 4}
\end{figure}

\begin{figure}
\centering
\includegraphics[width=12cm]{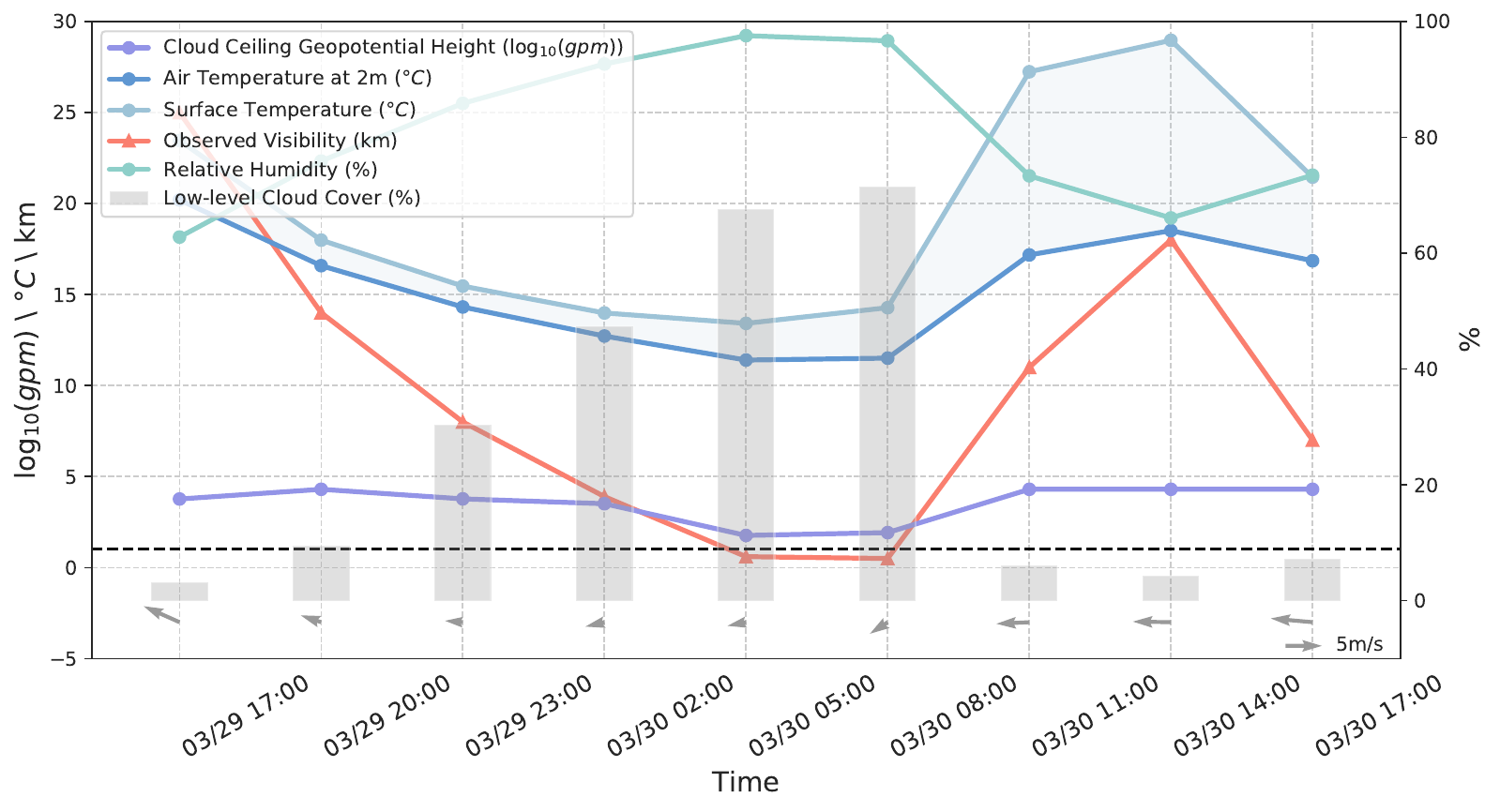}
\caption{Observed visibility and six meteorological variables at Station 111 from 17:00 on March 29th to 17:00 on March 30th, 2018. The shaded area indicates the difference between the air temperature at 2 meters and surface temperature. The black horizontal dashed line indicates a visibility threshold of 1 km. The station’s location is shown in Figure \ref{Figure 1}.}
\label{Figure 5}
\end{figure}

We utilized a widespread sea fog event to provide evidence for the validity of our findings. The sea fog event occurred from 19:00 on March 29th to 08:00 on March 30th, 2018 and was observed at 9 stations (see Figure \ref{Figure 4}). Specifically, we plotted the time evolution of six meteorological variables for Station 111 (see Figure \ref{Figure 5}), all of which exhibit a strong correlation with visibility (see Figure \ref{Figure 2}). These variables include relative humidity, cloud ceiling geopotential height, low-level cloud cover, air temperature at 2 meters, surface temperature, and wind speed at 2 meters. Between 16:00 on March 29th and 07:00 on March 30th, the relative humidity gradually rose to 90\% due to decreased air temperature and the southeast wind carrying a significant amount of water vapor. The surface temperature also decreased during this period. When moist air comes in contact with a cold subsurface, the ground surface cools the water vapor in the air below the dew point, leading to the formation of fog. Throughout this period, the temperature difference between the atmosphere and sea surface remained at approximately -2 °C, causing water vapor to sublimate upward and mix with the colder air, resulting in the formation of fog. Starting at 16:00 on March 29th, the southeast wind slowly shifted to the east with decreasing wind speed, providing favorable conditions for fog formation and maintenance. Moreover, a low cloud ceiling geopotential height implies the presence of low clouds or fog. The increase in wind speed from 7:00 on March 30th suggested that the lower atmosphere was becoming unstable. Simultaneously, the air and sea surface temperatures began to increase, leading to a decrease in relative humidity. All these factors contributed to the dissipation of the fog.

\subsection{Forecasting Methods}

As ground truth for model training, we separated sea fog events into two groups: fog (visibility $\le$ 1 km) and fog-free (visibility $>$ 1 km). The predictors of the forecasting model can be split into five categories: 1) time-lagged predictors extracted using the TLCA method; 2) station location (in terms of latitude and longitude); 3) hour, day, and month; 4) visibility observations for the 6 hours preceding the WRF-NMM launch; and 5) forecast lead time (from 1 to 60 hours). We used LightGBM models \cite{ke2017a}. LightGBM is a gradient-boosted tree-based model that is popular in the machine learning community for its high modeling quality, rapid parallel computing, and intuitive interpretability, to make forecasts for different types of sea fog.

The focal loss and ensemble learning are used to improve the ML model's prediction abilities while addressing the issue of data imbalance between fog and fog-free events, as shown in Figure \ref{Figure 1}(b). Object detection training tasks often suffer from uneven data distributions, prompting researchers to develop the focal loss function to address this issue \cite{lin2018a}. To address the class imbalance problem, the focal loss function employs a weighting parameter $\alpha$ to fix the ratio of positive and negative samples (see Eq. 2), where y and $y^\prime$ refer to the ground truth and predicted probability, respectively. The focusing parameter ($ \gamma > 0$) can be adjusted to minimize the relative loss for well-classified examples while placing extra emphasis on misclassified examples. Notably, when $\gamma$ is set to 0, the loss function becomes equivalent to the standard cross-entropy loss.

\begin{linenomath*}
\begin{equation}
L_{f l}=\left\{\begin{array}{cc}-\alpha\left(1-y^{\prime}\right)^\gamma \log y^{\prime}, \quad & y=1 \\ -(1-\alpha) y^{\prime} \log \left(1-y^{\prime}\right), & y=0\end{array}\right.
\label{Eq 2}
\end{equation}
\end{linenomath*}

\begin{figure}
\centering
\includegraphics[width=12cm]{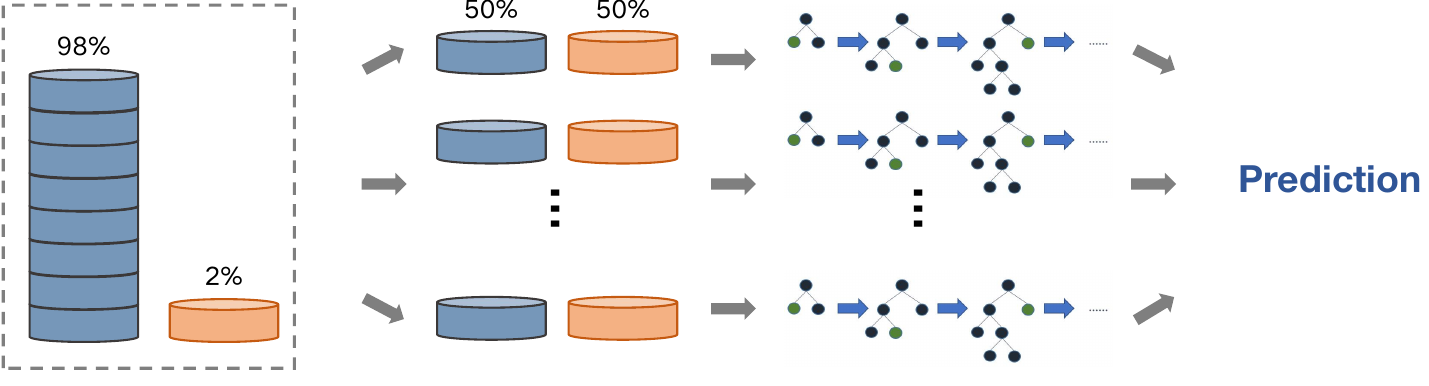}
\caption{The framework of our forecasting method based on ensemble learning and data resampling.}
\label{Figure 6}
\end{figure}

Ensemble learning combines predictions from multiple models to produce a single result, thereby enhancing the prediction accuracy by integrating outcomes across diverse models \cite{galar2012a}. In our study, we employed ensemble learning to train 10 LightGBM models (as illustrated in Figure \ref{Figure 6}). Before training, we divided the original, imbalanced dataset into N distinct halves and ensured a consistent ratio of fog and fog-free events by randomly resampling fog samples within each subset. Then, we trained a LightGBM model independently for each subset. To generate our final forecast, we used an average of the outputs of N distinct forecasting models.

\section{Experiments}

The dataset is split chronologically into three subsets: training, validation, and testing datasets. The training dataset comprises four years of data from 2014 to 2017, while the validation dataset is from 2018. The out-of-sample testing dataset spans from 2019 to 2020. In this study, sea fog prediction is treated as a binary event with two possible outcomes (yes or no, i.e., 1 or 0, respectively). A confusion matrix showing all possible outcomes is presented in Table \ref{Table 1}. The evaluation metrics used in our study comprise the probability of detection (POD), false alarm ratio (FAR), and equitable threat score (ETS), as described in Table \ref{Table 2}. The ETS has been widely used to evaluate fog forecasts in previous studies \cite{rom2016a, wang2014a} and is a common measure for binary forecasts \cite{jolliffe2012a}. Nonetheless, prior research has indicated that the expected ETS of a random forecasting system drops below 0.01 only when the number of samples $n > 30$ \cite{hogan2010a}. Hence, we employed the Heidke skill score (HSS), which is equitable for all values of $n$. Apart from the data used for training, we set aside an independent dataset from 2019 to 2020 to evaluate the model's performance and compared it against that of the WRF-NMM predictions and the NOAA FSL method (see Appendix B, NOAA FSL method).

\begin{table}[h]
\centering
\caption{Confusion matrix for calculating scalar performance metrics.}
\label{Table 1}
\begin{tabular}{llll}
\hline
\multirow{2}{*}{\textbf{Forecast}} & \multicolumn{3}{c}{\textbf{Observation}} \\ \cline{2-4} 
    &   Fog   &  Fog-free   &  Total  \\ \cline{2-4} 
    Fog          &   Hits (a)    &   False alarm (b)    &   a+b   \\ 
    Fog-free         &  Misses (c)     &  Correct Rejections (d)      &  c+d    \\ 
    Total   &   a+c    &   b+d    &  a+b+c+d   \\ \hline
\end{tabular}
\end{table}

\begin{table}[]
\newcommand{\tabincell}[2]{\begin{tabular}{@{}#1@{}}#2\end{tabular}}  
\centering
\caption{Evaluation metrics used in our analysis.}
\label{Table 2}
\begin{tabular}{lll}
\hline
 \textbf{Performance metric} & \textbf{Symbol} & \textbf{Equation} \\ \hline
 Probability of Detection & POD & $a/\left(a+c\right)$ \\ \hline
 False Alarm Ratio & FAR & $b/\left(a+d\right)$ \\ \hline
 Equitable Threat Score & ETS  &  \tabincell{l}{$\left(a-a_r\right) /\left(a+b+c-a_r\right)$ \\ $ a_r=(a+b)(a+c) / n $} \\ \hline
 Heidke skill score & HSS & \tabincell{l}{$\mathrm{HSS}=\frac{\mathrm{PCF}-E}{1-E}$ \\ $\mathrm{PCF}=\frac{(a+d)}{n}$ \\ $E=\left(\frac{a+c}{n}\right)\left(\frac{a+b}{n}\right)+\left(\frac{b+d}{n}\right)\left(\frac{c+d}{n}\right)$} \\ \hline
\end{tabular}
\end{table}

\section{Results}

\begin{figure}
\centering
\includegraphics[width=12cm]{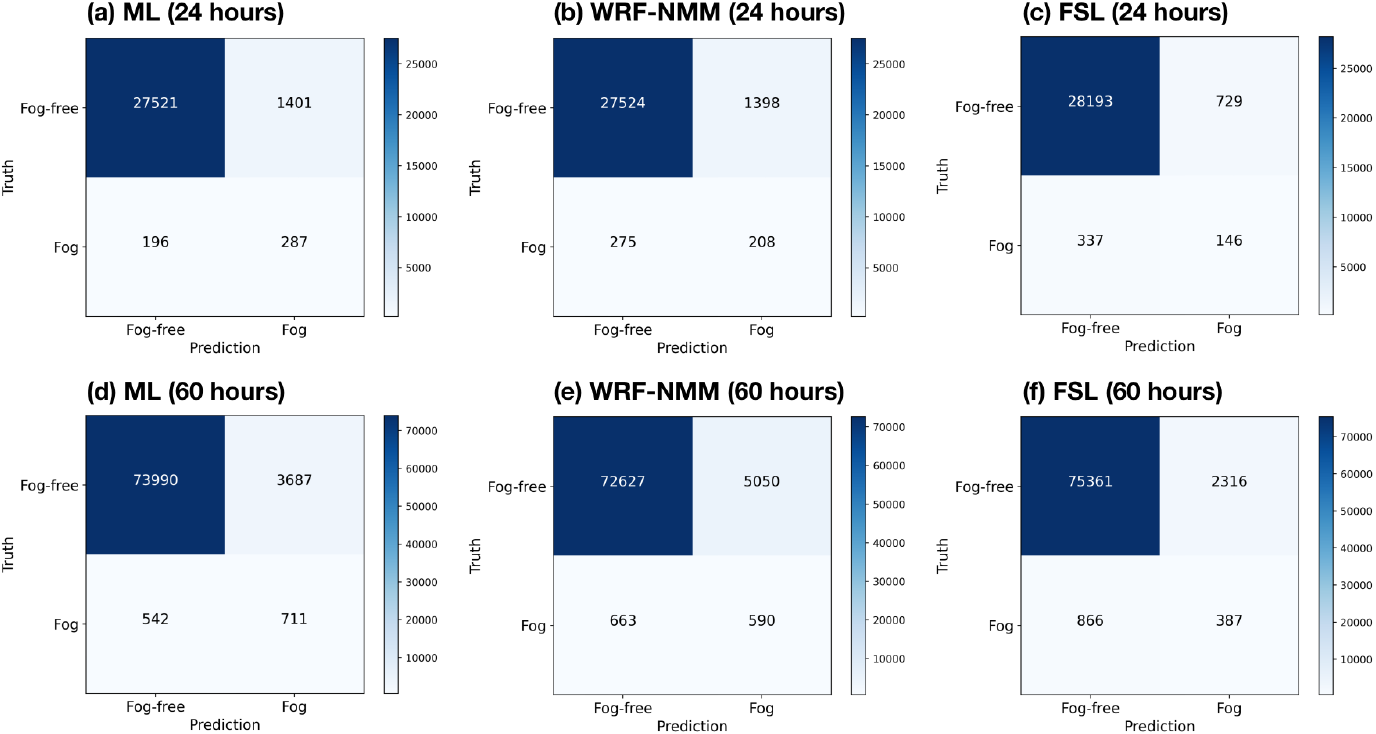}
\caption{Confusion matrices of the forecasting abilities of 3 methods, including our ML-based method (a, d), the WRF-NMM (b, e), and FSL algorithm (c, f). (a-c) 24-hour lead time; (d-f) 60-hour lead time.}
\label{Figure 7}
\end{figure}

\begin{figure}
\centering
\includegraphics[width=12cm]{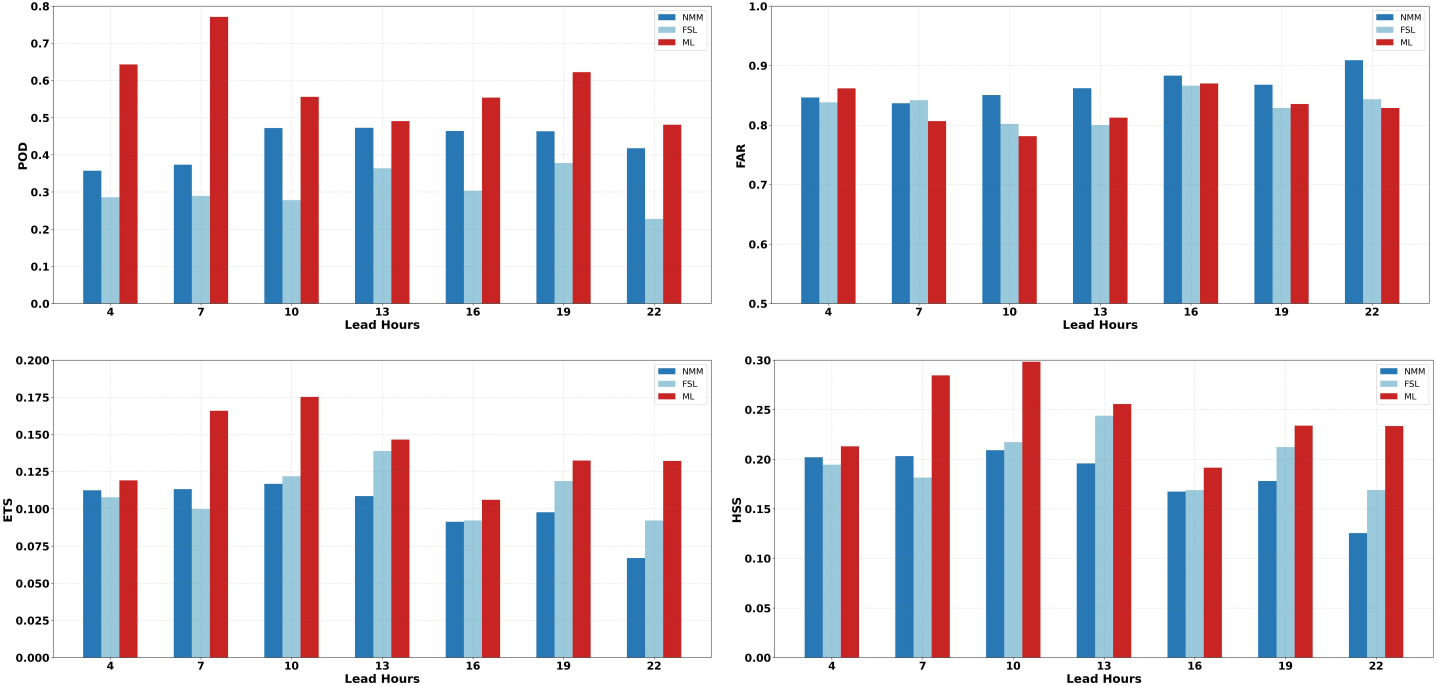}
\caption{The forecasting abilities (POD, FAR, ETS, and HSS) in terms of the 24-hour lead time of the 3 methods, namely, the ML method, the WRF-NMM, and the FSL algorithm.}
\label{Figure 8}
\end{figure}

\begin{figure}
\centering
\includegraphics[width=12cm]{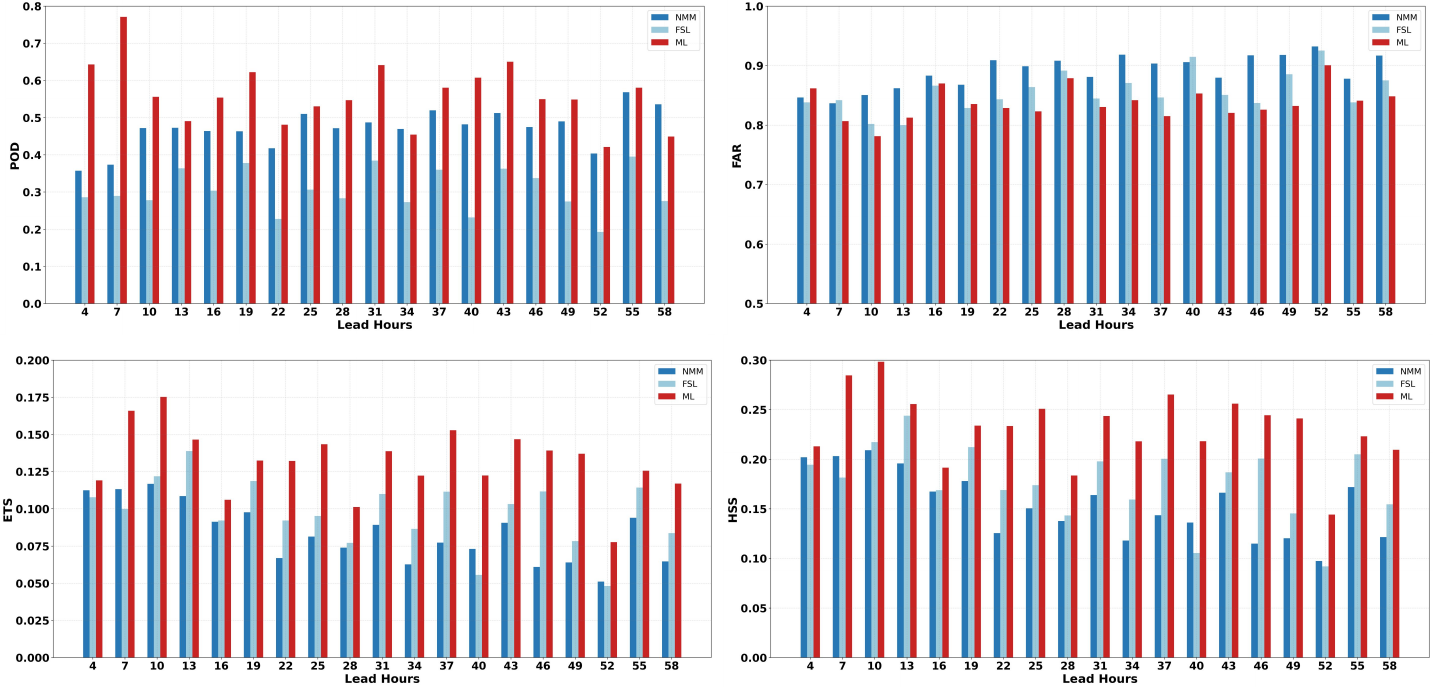}
\caption{The forecasting abilities (POD, FAR, ETS, and HSS) in terms of the 60-hour lead time of the 3 methods, namely, the ML method, the WRF-NMM, and the FSL algorithm.}
\label{Figure 9}
\end{figure}

\textbf{Our ML-based models have demonstrated comparable performance in predicting sea fog with a horizontal visibility of 1 km.} Our ML-based method outperformed the WRF-NMM and FSL algorithms in terms of the average values of the four metrics (POD, FAR, ETS, and FAR) for both the 24-hour and 60-hour lead times (see Table \ref{Table 3} and Figure \ref{Figure 7}). Notably, this improvement was achieved by enhancing the probability of detection (POD) while reducing the false alarm ratio (FAR). Specifically, for the 3-hour interval forecast (see Figures \ref{Figure 8} and \ref{Figure 9}), our ML-based method exhibited an obvious advantage over the other two methods.

\begin{table}[]
\centering
\caption{Averaged forecasting abilities (POD, FAR, ETS, and HSS) for 24-hour and 60-hour lead times.}
\label{Table 3}
\begin{tabular}{ccccccc}
\hline
\multirow{2}{*}{\textbf{Metric}} & \multicolumn{3}{c}{\textbf{24-hours}} & \multicolumn{3}{c}{\textbf{60-hours}} \\ \cline{2-7} 
    ~ & \textbf{ML}  &  \textbf{WRF-NMM}  &   \textbf{FSL}  &  \textbf{ML}  &   \textbf{WRF-NMM}  &  \textbf{FSL} \\ \hline
    \textbf{POD}  &  \textbf{0.5393}   &    0.4316  &   0.3037  &  0.5232  &  0.4709    &  0.3054  \\ \hline
    \textbf{FAR}  &  \textbf{0.8208}   &   0.8649  &  0.8315  &  0.8301  &   0.89  &   0.856  \\ \hline 
    \textbf{ETS}  &   \textbf{0.1426}  &   0.101  &  0.1102  &  0.135  &  0.0837  &  0.0972  \\ \hline
    \textbf{HSS}  &  \textbf{0.2491} &  0.183    &  0.1981  &   0.2374   &   0.1539   &  0.1764  \\ \hline
\end{tabular}
\end{table}

\textbf{The TLCA method improved the prediction accuracy by selecting optimal predictors.} We compared the predictors identified by the TLCA method with two other sets of predictors, one including all variables from WRF-NMM outputs without time lags and the other including all variables from WRF-NMM outputs with a 6-hour time lag. The experiments were conducted with identical configurations. Table \ref{Table 4} presents the evaluation metrics (POD, FAR, ETS, and HSS) in terms of a 60-hour lead time. The TLCA-extracted predictors enhanced the forecasting performance by identifying factors that exhibit strong correlations with fog. Furthermore, despite the reduction in the number of predictors, the forecasting abilities can still be improved.

\textbf{The utilization of the focal loss function provides guidance to the model, enabling it to prioritize fog events and alleviate challenges arising from severe data imbalance.} We conducted an evaluation of the impact of different loss functions on the ML model, including the conventional cross-entropy loss and the focal loss. Two sets of parameters were employed: $\alpha=0.2$, $\gamma=2$ and $\alpha=0.2$, $\gamma=4$. The outcomes presented in Table \ref{Table 4} demonstrate that the utilization of focal loss functions improves the POD, albeit with a slight increase in the FAR. Moreover, based on the ETS and HSS metrics, it is evident that the focal loss functions yield superior forecasting abilities compared to those of the standard cross-entropy loss. Consequently, incorporating the focal loss function during the training process enables the ML model to concentrate more effectively on fog events, thus enhancing its capability to forecast such events.

\textbf{Forecasting abilities may potentially be enhanced by combining ensemble learning with a data resampling strategy.} In our experiments, we compared five learning strategies: direct model training (without Easy-ensemble), Easy-ensemble with undersampling, Easy-ensemble with oversampling, undersampling alone, and oversampling alone. The resampling ratios for both undersampling and oversampling were set to 0.1. The parameters of the focal loss were set to $\alpha=0.2$ and $\gamma=4$. The comparison results presented in Table \ref{Table 4} illustrate that utilizing Easy-ensemble with either undersampling or oversampling leads to improved forecasting abilities, as evidenced by higher scores in terms of the ETS and HSS. This improvement can be attributed to a simultaneous reduction in both the POD and FAR.

The results of this study highlight that integrating an ML model with an NWP model output can significantly improve the accuracy of fog forecasts. However, a notable limitation encountered was the data imbalance issue between fog and fog-free samples within the training dataset, which adversely impacted the performance of the ML model. To address this challenge, we employed the focal loss function and the Easy-ensemble learning strategy, which effectively mitigated the data imbalance problem and yielded significant enhancements in forecasting abilities.

\begin{table}[]
\centering
\caption{Forecasting abilities of varied predictors, loss functions, and learning strategies.}
\label{Table 4}
\begin{tabular}{cccccc}
\hline
    &  & \textbf{POD} & \textbf{FAR} & \textbf{ETS} & \textbf{HSS} \\ \hline
\multirow{3}{*}{Predictor} & All variables without time-lagged & 0.119 & 0.699 & 0.089 & 0.16 \\
    & All variables with time-lagged & 0.126 & 0.661 & 0.097 & 0.18 \\
    & Variables from TLCA & 0.139 & 0.695 & 0.101 & 0.18 \\ \hline
\multirow{3}{*}{Loss function} & Cross-Entropy & 0.139 & 0.695 & 0.101 & 0.18 \\
    & Focal Loss (0.2, 2) & 0.638 & 0.858 & 0.119 & 0.21  \\
    & Focal Loss (0.2, 4) & 0.639 & 0.858 & 0.118 & 0.21 \\ \hline
\multirow{5}{*}{Learning strategy} & Without Easy-ensemble & 0.56 & 0.842 & 0.128 & 0.23 \\
    & Easy-ensemble \& under-sample & 0.567 & 0.838 & 0.132 & 0.23 \\
    & Easy-ensemble \& over-sample & 0.548 & 0.84 & 0.129 & 0.23 \\
    & Under-sample & 0.559 & 0.838 & 0.131 & 0.23 \\
    & Over-sample & 0.555 & 0.842 & 0.128 & 0.23 \\ \hline
\end{tabular}
\end{table}

\section{Conclusions}

We have developed a novel ML-based method for predicting sea fog using the Yangtze River Estuary (YRE) coastal area as an illustration. We focus on a visibility of 1 km and a lead time of up to 60 hours. To accomplish this, we have introduced a technique named time-lagged correlation analysis (TLCA). By utilizing TLCA, we can identify crucial predictors with time-lagged characteristics and delve into the underlying mechanisms that give rise to sea fog. To address the challenge of imbalanced data between fog and fog-free samples encountered during the training of our model, we employed both the focal loss and ensemble learning strategies.

In addition, we have constructed a unified ML-based forecasting method capable of being applied across different stations and seasons. When tested against two traditional methods, namely, the WRF-NMM and the NOAA FSL method, our ML-based model demonstrates superior performance in predicting sea fog with a lead time of 60 hours and visibility limited to 1 kilometer. Specifically, our model not only enhances the probability of detection (POD) of sea fog but also reduces the false alarm ratio (FAR).

However, it should be noted that further research is required to refine and enhance the forecast method presented in this study. While our current method generates classification forecasts distinguishing between visibility of 1 kilometer or less and visibility exceeding that threshold, we aspire to predict actual visibility values and inclement weather associated with poor visibility in the future. To refine our methodology, it is imperative to consider spatial correlations among meteorological variables when identifying predictors. Additionally, we can explore the utilization of deep learning models equipped with convolutional structures to automatically extract features from fields of meteorological variables. This advancement would enable us to capture a more comprehensive representation of the weather situation at hand.

\appendix

\section{Variables from the WRF-NMM model}

In this study, historical forecast data from the weather research and forecasting nonhydrostatic mesoscale model (WRF-NMM) were provided by the US National Centers for Environmental Prediction (NCEP) \cite{janjic2003a}. The dataset features a horizontal resolution of $8 \times 8$ kilometers  and encompasses the entire Yangtze River Estuary (YRE) region. The WRF-NMM produces hourly predictions for a 60-hour forecast period, launching twice daily at 00:00 UTC and 12:00 UTC. The forecast variables incorporated in the model include temperature, relative humidity, cloud cover, wind speed, precipitation, and more. We selected the 26 variables listed below and omitted certain nearly constant variables.

\begin{table}[h]
\caption{Description of the 26 variables selected from the WRF-NMM.}
\label{Table A1}
\begin{tabular}{p{4cm} p{4cm} p{5cm}}
\hline
\textbf{Variables} & \textbf{Long Name} & \textbf{Description} \\ \hline
DPT GDS3 HTGL & Dew point & Height in 2 meters \\ 
HGT\_GDS3\_0DEG &	Geopotential height & Level of 0 deg (C) isotherm \\
HGT\_GDS3\_CEIL & Geopotential height	& Cloud ceiling \\
HGT\_GDS3\_HTFL & Geopotential height	& Highest tropospheric freezing level \\
HGT\_GDS3\_SFC	& Geopotential height	& Ground/water surface \\
T\_CDC\_GDS3\_EATM	& Total cloud cover	& Entire atmosphere \\
H\_CDC\_GDS3\_HCY	& High cloud cover	& High cloud layer \\
M\_CDC\_GDS3\_MCY	& Mid cloud cover	& Mid cloud layer \\
L\_CDC\_GDS3\_MCY	& Low cloud cover	& Low cloud layer \\
PLI\_GDS3\_SPDY	& Parcel lifted index (to 500 hPa)	& Layer between two levels at specified pressure difference from ground to level \\
LFT\_X\_GDS3\_ISBY	& Surface lifted index	& Layer between two isobaric levels \\
PRES\_GDS3\_SFC	& Pressure	& Ground/water surface \\
PRMSL\_GDS3\_MSL	& Pressure reduced to Mean sea level & 	Mean sea level \\
MSLET\_GDS3\_MSL	& Mean sea level pressure	& Mean sea level \\
P\_WAT\_GDS3\_EATM	& Precipitable water &	Entire atmosphere \\
POP\_GDS3\_SFC	& Probability of precipitation	& Ground/water surface \\
R\_H\_GDS3\_HTGL	& Relative humidity	& Height in 2 meters \\
R\_H\_GDS3\_HYBL	& Relative humidity	& Hybrid level \\
SPF\_H\_GDS3\_HTGL	& Specific humidity	& Height in 2 meters \\
SPF\_H\_GDS3\_SPDY	& Specific humidity	& Layer between two levels at specified pressure difference from ground to level \\
TMP\_GDS3\_HTGL	& Temperature	& Height in 2 meters \\
TMP\_GDS3\_SFC	& Temperature	& Ground/water surface \\
U\_GRD\_GDS3\_HTGL &	U-component of wind	& Height in 10 meters \\
U\_GRD\_GDS3\_SPDY	& U-component of wind	& Layer between two levels at specified pressure difference from ground to level \\
V\_GRD\_GDS3\_HTGL	& V-component of wind &	Height in 10 meters \\
V\_GRD\_GDS3\_SPDY	& V-component of wind &	Layer between two levels at specified pressure difference from ground to level \\ \hline
\end{tabular}
\end{table}

\section{NOAA FSL method}

The US National Oceanic and Atmospheric Administration (NOAA) Forecast Systems Laboratory (FSL) method \cite{doran1999a}: $VIS_{FSL} = 1.609 \times 6000 \times (T - T_d) /rh^{1.75}$, where $rh$ is the relative humidity and $T – T_d$ is the dew point depression ($^\circ C$).

\section*{Data Availability Statement}

The LightGBM is a public model, and its releases are accessible through this website: https://lightgbm.readthedocs.io/en/stable/index.html. 
Due to the nature of this research, participants in this study did not agree for their data to be shared publicly, so supporting data are not available.



\acknowledgments

This work is supported by the National Key Research and Development Program of China (2022YFE0195900, 2021YFC3101600, 2020YFA0608000, 2020YFA0607900), the National Natural Science Foundation of China (42125503, 42075137) and the Science and Technology Plan Project of Ningbo (2022S181).

%
%
\bibliography{August-2022-latex-templates/cite}

%
%
%
%
%

\end{document}